\documentclass[letterpaper, 10 pt, conference]{ieeeconf}  % Comment this line out if you need a4paper

\IEEEoverridecommandlockouts                              % This command is only needed if 
\usepackage{amsmath} % assumes amsmath package installed
\usepackage{amssymb}  % assumes amsmath package installed

\title{\LARGE \bf Decentralized Circle Formation Control for Fish-like Robots in the Real-world via Reinforcement Learning*
}

\author{Tianhao~Zhang$^{1}$$^{\dag}$, Yueheng~Li$^{1}$$^{\dag}$, Shuai~Li$^{1}$, Qiwei~Ye$^{2}$, Chen~Wang$^{1,3}$, and Guangming~Xie$^{1}$% <-this % stops a space
\thanks{*This work was supported in part by grants from the National Natural Science Foundation of China (NSFC, No.61973007, 61633002). \emph{Corresponding author: C. Wang (wangchen@pku.edu.cn})}
\thanks{$^{1}$The State Key Laboratory of Turbulence and Complex Systems, Intelligent Biomimetic Design Lab,
       College of Engineering, Peking University, Beijing 100871, China.}
\thanks{$^{2}$Microsoft Research Asia, Beijing 100871, China.}
\thanks{$^{3}$National Engineering Research Center of Software Engineering, Peking University, Beijing 100871, China.}
\thanks{$^{\dag}$First two authors contributed equally to this work.}%
}

\usepackage{graphicx}
\usepackage{color}
\usepackage{amsmath}
\usepackage{mathtools}
\usepackage{amsthm,amssymb}
\usepackage{mathrsfs}
\usepackage{booktabs}

\newcommand{\tabincell}[2]{\begin{tabular}{@{}#1@{}}#2\end{tabular}}

\begin{document}

\maketitle
\thispagestyle{empty}
\pagestyle{empty}

%%%%%%%%%%%%%%%%%%%%%%%%%%%%%%%%%%%%%%%%%%%%%%%%%%%%%%%%%%%%%%%%%%%%%%%%%%%%%%%%
\begin{abstract}
In this paper, the circle formation control problem is addressed for a group of cooperative underactuated fish-like robots involving unknown nonlinear dynamics and disturbances.
Based on the reinforcement learning and cognitive consistency theory, we propose a decentralized controller without the knowledge of the dynamics of the fish-like robots. 
The proposed controller can be transferred from simulation to reality. It is only trained in our established simulation environment, and the trained controller can be deployed to real robots without any manual tuning.
Simulation results confirm that the proposed model-free robust formation control method is scalable with respect to the group size of the robots and outperforms other representative RL algorithms.
Several experiments in the real world verify the effectiveness of our RL-based approach for circle formation control.

\end{abstract}

%%%%%%%%%%%%%%%%%%%%%%%%%%%%%%%%%%%%%%%%%%%%%%%%%%%%%%%%%%%%%%%%%%%%%%%%%%%%%%%%
\section{INTRODUCTION} \label{sec:intro}
%

%In nature, most fishes swim in a group to form a circle formation (see Fig. \ref{fig:Torus}A) during migration and predation \cite{vicsek2012collective}, which has attracted biologists, physicists, and engineers to explore the mechanisms.
% %
% This fascinating phenomenon has long been hypothesized that schooling fish may be able to obtain hydrodynamic benefits by swimming closely with others \cite{breder1965vortices}.
% %
% Excitingly, with the aid of fish-like robots, Li {et al.} demonstrate that fish could save energy or generate thrust by exploiting neighbor-induced vortices in water, and swimming to form the circle may be the best energy-saving for fish schools since each fish could exploit neighbor-induced vortices \cite{Li2020FishVortex}.
%

\begin{figure}[t]
    \centering
    \includegraphics[width=0.9\linewidth]{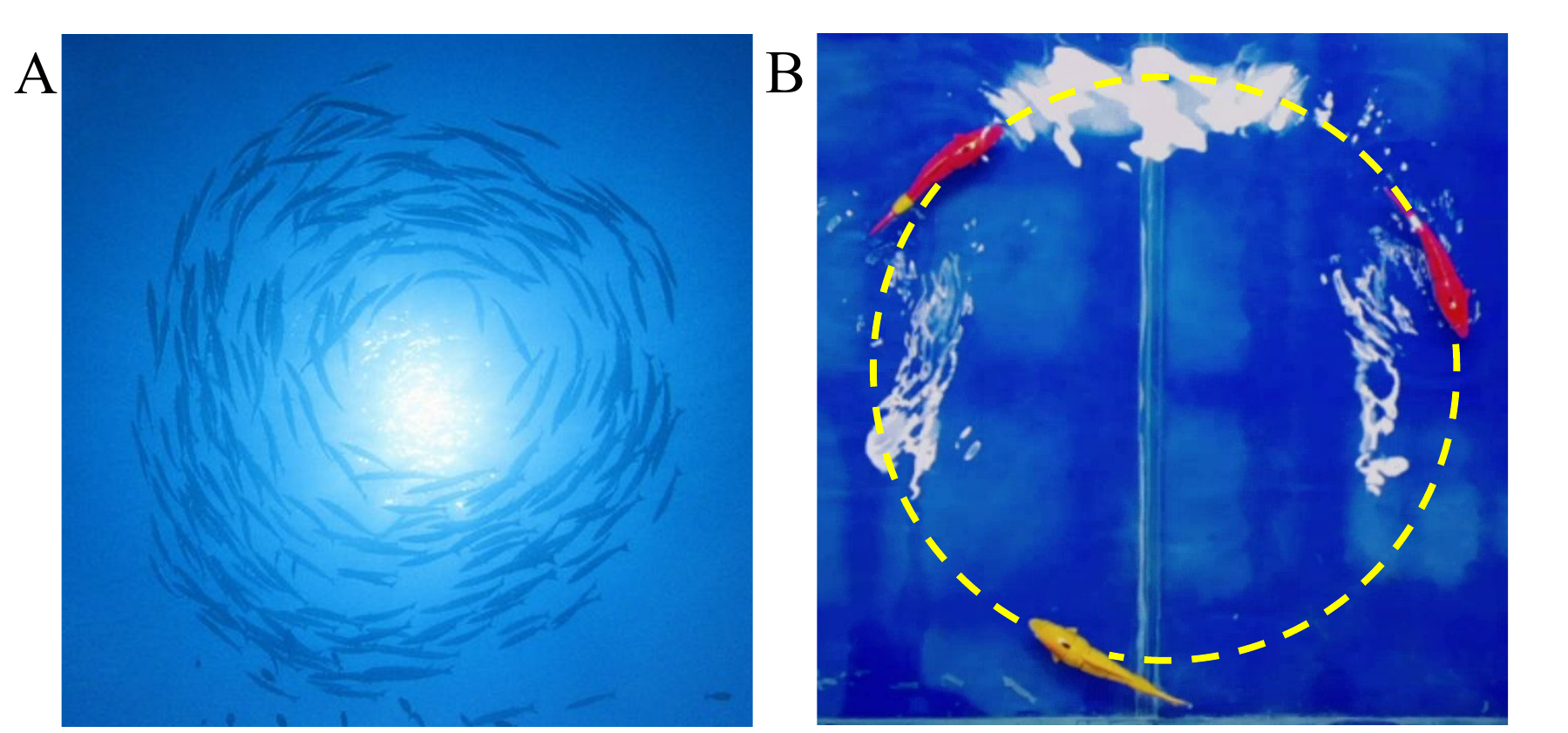}
    \caption{Torus formation of fish schools in nature (A), and the corresponding schematics of fish-like robots swarm in physical experiments (B).\footnotesize{ (A) is {adopt from {https://www.pinterest.com/pin/358669557799034652/}}}}
    \label{fig:Torus}
\end{figure}

In nature, many fish species live in groups, and can get benefits of group formation for such as foraging, keeping warm, reducing the cost of migration \cite{brown2011fish,krause2002living,Li2020FishVortex}.
\textit{Torus} (Fig. \ref{fig:Torus}A), one of the most common formations of fish groups, is especially beneficial when resting or defending against predators \cite{vicsek2012collective}.
Inspired by the torus formation behavior of real fish, circle formation control of fish-like robots (Fig. \ref{fig:Torus}B) has been an emerging research topic, and it has practical potential in various complex tasks, such as marine exploration and rescue.
However, the fish-like robot is an underactuated system subjected to highly nonlinear dynamics.
Meanwhile, the fish-like robots swim in water via deforming bodies, and the waves caused by that have a dramatic interference in their motion \cite{WaXiWaCa11,zhang2020path}.
Therefore, under these disturbances and the influence of high nonlinearities, it is challenging to design a robust controller for fish-like robots to achieve desired circle formations. 
%

%传统控制方法
Much effort has been devoted to theoretical studies on circle formation control in the last decade, some of which are reviewed here.
Considering the dynamic models of the robots, there are two major categories of the research, one is for the mass point model and the other is for the unicycle model. 
For the mass point model, it is relatively easy to design a law to drive robots to a circle. 
A kind of typical studies is \cite{wang2017limit,wang2019limit}, they use the limit-circle based control laws to drive a group of points to form a prescribed circle. 
In \cite{Li2018a}, a localization and circumnavigation method is proposed for particles in three-dimensional space. 
For the nonholonomic unicycle model, there are also many works.
A distributed control law using only local measurement is proposed in \cite{Zheng2015}, which can drive unicycle-like robots to encircle a target with different orbits. Unlike \cite{Zheng2015} using relative positions, \cite{Yuxiao2018} presents a law that uses only bearing measurement and can achieve circle formation with an identical radius. 
Taking multiple targets into account, \cite{Shi2020} further designs a control law that can drive vehicles to circle different targets. 
Although these theoretical control laws perform well in simulation and can be provably guaranteed to converge and stabilize, they are rarely implemented to real robots like unmanned aerial vehicles (UAVs) or unmanned ground vehicles (UGVs). 
This is mainly because these laws depend on continuous state and action space, as well as time horizon, which is impractical in the real-world. 
Moreover, they can hardly deal with the unpredictable and inevitable disturbance and noise when applied to real robots or vehicles.
Considering that fish-like robots suffer from high nonlinearities and disturbances, these traditional methods may work little in fish-like robots.
%
% Therefore, group swimming using robotic platforms has scarcely been studied.

To the best of our knowledge, there are only two works that use physical fish-like robots for the circle formation control experiments \cite{li2019bottom,zhang2020global}.
In \cite{li2019bottom}, by combining active disturbance rejection control and proportional navigation law, the bottom-level motion controller of a single robot is proposed, which includes the target position control and target pose control.
Then, with the aid of the motion controller, each of the three fish-like robots swims towards its individual virtual target position and posture to formulate the circle formation in a plane.
Similarly, in \cite{zhang2020global}, through the position and posture control of the singe agent by the proportional controller, three soft fish-like robots swim on a common circle in a plane.
However, both of them consider the circle formation control of fish-like robots through the bottom-level motion control of a single robot, thus they both emphasize the design of the robust dynamic model of their specific robots rather than the circle formation methods.
Considering that different robots have different motion controllers, these kind of model-based methods may not universal to other robots.
Therefore, compare with these model-based methods, a circle formation control approach that does not rely on the specific dynamic model is more appealing.

%% 强化学习
Recently, reinforcement learning (RL)-based methods have generated excitement in the robotic control system because their approach to obtaining strategies by trial and error is independent of the robotic dynamic model.
Using the RL-based methods, the trajectory tracking \cite{cui2017adaptive} and the depth control \cite{wu2019depth} for a single AUV are achieved in the simulation.
However, like the above two works, most of the RL-based control research is limited to computer simulation.
One reason is that, if learning strategies through real-world experiments, the trial and error mechanism of RL is not only time-consuming but also deleterious to robots.
For another, Boeing {et al.} \cite{boeing2012leveraging} indicated that different high-quality physics engines can give radically different predictions for a single problem, which shows the great challenge of transferring a control system from a fluid simulation to a real robot.
Recently, some works have demonstrated the real-world application of model-free RL-based control of a single underwater robot.
In \cite{yu2020underwater}, the DDPG RL method is utilized for the target tracking control of a fish-like robot.
In \cite{zhang2020path}, the path-following task of the fish-like robot is successfully realized in the physical environment based on the advantage actor-critic (A2C) RL method.
However, both two works are about the control of a single fish-like robot.
Besides, the single-agent reinforcement learning methods they used (i.e., DDPG and A2C) focusing on maximizing the cumulative reward for a single agent may be not suitable for multi-agent tasks \cite{hernandez2019survey}.
%

%% 本文工作
To sum up, it is still an open research problem to design a model-free robust circle formation controller for groups of underactuated fish-like robots subjected to high nonlinearities and disturbances in the fluid.
To address this problem, in this paper, we propose an RL-based learning approach to get a robust model-free controller for the circle formation of fish-like robots.
The proposed RL-based approach works in a decentralized manner that the controller of each robot does not rely on the global information during execution. Thus, it is scalable with respect to the group size of the robots.
% Considering that decentralized control is more appealing for multi-agent systems than centralized control due to its scalability and \todo{robustness with agent failures}, our proposed RL-based controller is decentralized, which means the controller of each robot does not rely on the global information during execution.
%

Specifically, this paper first introduces a simulation environment based on the experimental motion data of the fish-like robot instead of its complex dynamic model, and gives a description of the circle formation control task.
Then, a novel cognitive consistency-based multi-agent RL (MARL) algorithm is proposed.
Our proposed MARL algorithm learns a robust circle formation controller for each agent by centralized training in the data-based simulation environment under the description of the circle formation control task.
Finally, the RL-based circle formation controller for the agent is directly deployed on the physical fish-like robot.
Without any further tuning or training, the formation controller for each agent executed based on its local observation can steer the group of physical fish-like robots to swim on a circle in planar.
%描述实验
Some experiments are conducted in the real world with three fish-like robots. Robots controlled by the proposed controller can accurately swim around the target with the preset radius and switch their distribution on the circle according to the precise control of the distance between each other.
The performance of the proposed RL method outperforms the traditional method and the representative RL baselines both in the simulation and real-world.

Compared with previous studies, this paper has three main features.
\begin{enumerate}
    \item It is the first time that an RL-based circle formation control approach is proposed for multiple fish-like robots. The proposed approach is in an end-to-end manner that can realize circle formation control without the position and pose controllers.
    \item By using the experimental motion data of the fish-like robot instead of its complex dynamic model, a data-driven based simulation environment is established, which solves the sim-to-real transfer problem. That is, the established simulation environment enables the RL strategies trained in it to be directly transferred to the physical environment without any further tuning.
    \item A novel cognitive consistency-based MARL algorithm is proposed. The simulation and real-world experiments show that such a decentralized execution algorithm has the potential to control a large-scale number of robots.
\end{enumerate}

\section{Preliminaries}

\subsection{Model and CTDE}\label{sec:CTDE}

For using reinforcement learning to model the cooperative multi-agent tasks, in this paper, the decentralized partially observable Markov decision processes (DEC-POMDP) \cite{oliehoek2016concise} is taken as the standard, as many previous works did.
% We take DEC-POMDP \cite{oliehoek2016concise} as the standard for modeling cooperative multi-agent tasks using reinforcement learning, as many previous works did.
%
A DEC-POMDP for $N$ agents is defined by a tuple $\mathcal{G}=\langle \mathcal{S},\mathcal{U},\mathcal{P},r,\mathcal{Z},\mathcal{O},\mathcal{N},\gamma\rangle$, where $\mathcal{S}$ is the set of global state $s$ of the environment. Each agent $i\in \mathcal{N}$ chooses an action $u_i\in \mathcal{U}$ at each time step, forming a joint action $u\in \mathcal{U}^N$. This causes a transition to the next state according to the state transition function $\mathcal{P}(s'|s,u):\mathcal{S}\times \mathcal{U}^N \times \mathcal{S}\to [0,1]$ and reward function $r(s,u):\mathcal{S}\times \mathcal{U}^N\to \mathcal{R}$ shared by all agents. $\gamma\in [0,1]$ is a discount factor. Each agent has individual, partial
observation $z\in \mathcal{Z}$ according to observation function $\mathcal{O}(s,i):\mathcal{S}\times \mathcal{N}\to \mathcal{Z}$. Each agent also has an action-observation history $\tau_i\in\mathcal{T}:(\mathcal{Z}\times \mathcal{U})^*$, on which it conditions a stochastic policy $\pi_i (u_i|\tau_i):\mathcal{T}\times U\to [0,1]$. The joint policy $\pi$ has a joint action-value function $Q^{\pi}(s_t,u_t)=\mathcal{E}_{s_{t+1:\infty},u_{t+1:\infty}}[\sum_{k=0}^{\infty}\gamma^k r_{t+k}|s_t,u_t]$.

For the decentralized control of cooperative MARL tasks, centralized training with decentralized execution (CTDE) is a common paradigm.  Through centralized training, the action-observation histories of all agents and the full state can be made accessible to all agents. These allow agents to learn and construct individual action-value functions correctly while selecting actions based on their own local action-observation history at execution time.

\subsection{Fish-like robot}\label{sec:CPG}

\begin{figure}[t]
    \centering
    \includegraphics[width=0.95\linewidth]{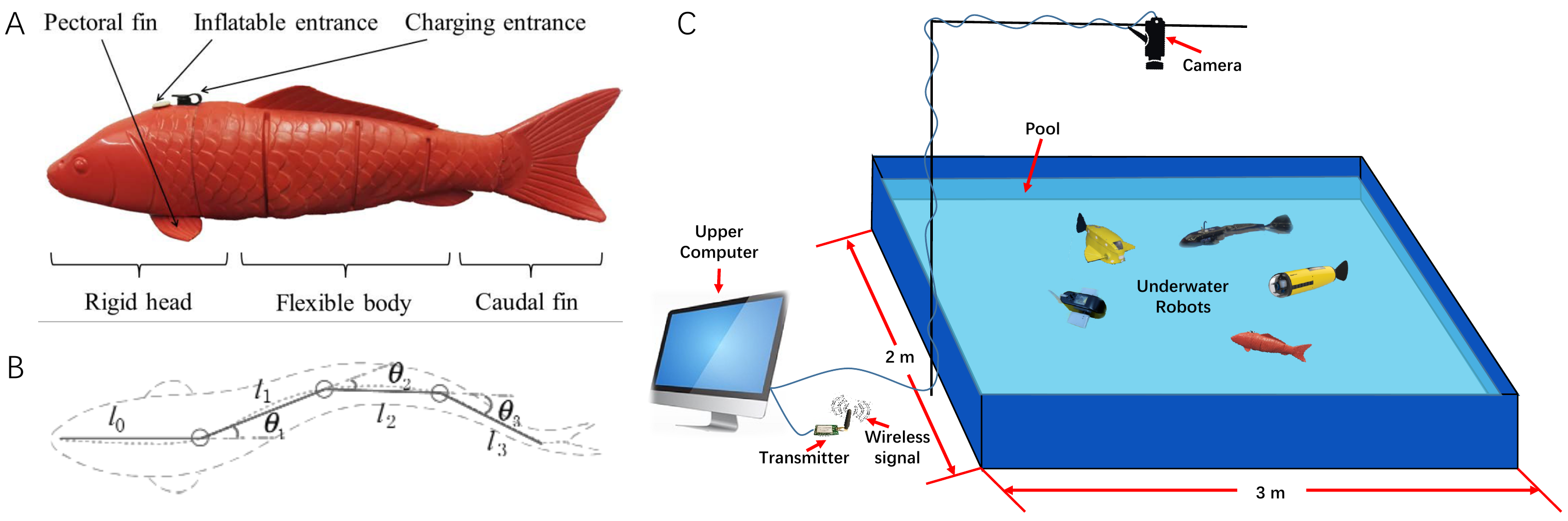}
    \caption{The prototype of the fish-like robot (A). The schematic of the three-joint propulsive structure (B). The physical platform consists of a $300 \times 200 cm$ pool, a server computer, an overhead camera, and a wireless communication module (C).}
    \label{fig:robot_and_platform}
\end{figure}

A typical widely concerned biomimetic fish-like robot \cite{zhang2020path} mimicking the Koi Carp is concerned in this paper.
%In this work, a kind of widely concerned fish-like robot \cite{zhang2020path} was chosen as the biomimetic underwater robot of interest.
The robot, with the $44.3 cm$ length and $0.85 kg$ weight, has a streamlined head, a flexible body, and a caudal fin (Fig. \ref{fig:robot_and_platform}A).
In the body, there are three joints linked together by aluminum exoskeletons, and each joint is driven by a servomotor (Fig. \ref{fig:robot_and_platform}B).
The robot swims just below the water surface since its density is close to water.
A typical center pattern generator (CPG) is utilized to control the three-joint servomotors for making the fish-like robot swim like the real fish.
The CPG model is as follows,
\begin{eqnarray}\label{eq:robot_CPG}
	\begin{cases}
		\dot{R}_i (t) = &\zeta_r ( \hat{R}_i-R_i(t) ) \\
		\dot{X}_i(t) = & \zeta_x ( \hat{X}_i - X_i(t)) \\
		\ddot{\Phi}_i(t) = & -\zeta_{\Phi}^2 \sum_{j=1, j\neq i}^{n} ( \Phi_i(t) - \Phi_j(t)  - \varphi_{ji})  \\ & - 2(n-1) \zeta_{\Phi} (\dot{\Phi}_i(t) - 2\pi f) \\
		\theta_i (t) = & x_i(t) + r_i(t) \sin (\Phi_i(t)), \quad i=1,2,\ldots, n
	\end{cases},
\end{eqnarray}
where $n=3$ is the number of the body joints, $\hat{R}_i$ and $\hat{X}_i$ are the desired swing amplitude and offset angle of the joint $i$, respectively, $\varphi_{ij}$ is the desired phase bias between joint $i$ and $j$, and $f$ is the desired swing frequency of each joint.
Three parameters ($\zeta_r, \zeta_x, \zeta_{\Phi}$) affect the related transient dynamics.
The output signal $\theta_i(t)$ represents the deflection angle of the corresponding joint $i$ at time $t$ (Fig. \ref{fig:robot_and_platform}B). 
More details about the robot and CPG refer to \cite{LiWaXi15}.

\section{DRL-based Approach design}

In this section, we propose a DRL-based approach to deal with the decentralized circle formation control task for multiple fish-like robots.
Some details will be described, including the formation control task, simulation environment establishment, cognitive consistency-based MARL method, and RL methods layout.

\subsection{Description of Circle Formation Task}\label{sec:CFT}

\begin{figure}[t]
    \centering
    \includegraphics[width=0.9\linewidth]{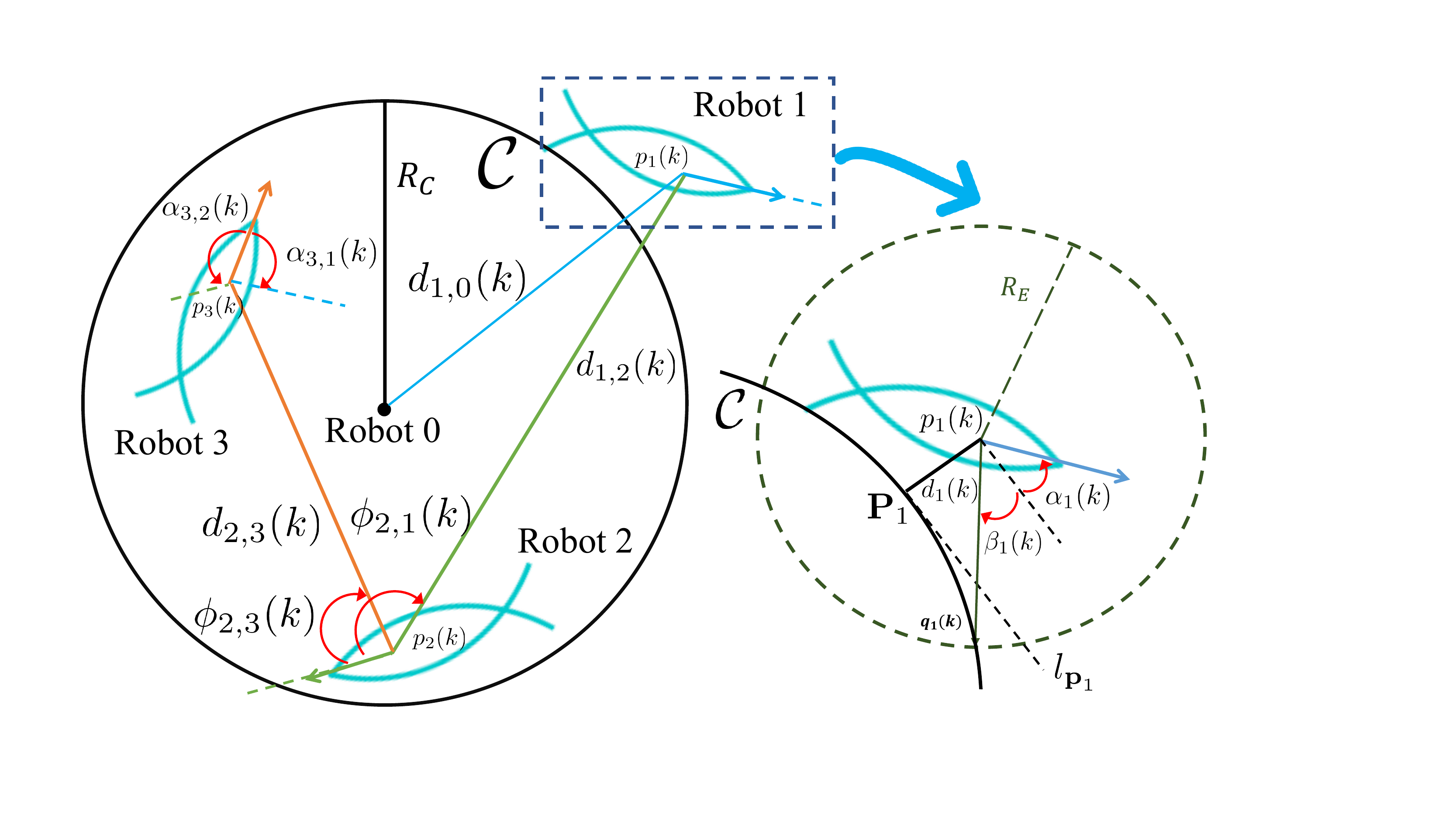}
    \caption{Circle formation task of $N$, $N\geq 2$, fish-like robots.% including the description of swimming around the target (shown on left) and the description of swimming on the path (shown on right).
    }
    \label{fig:formation_task}
\end{figure}

Like other similar works \cite{li2019bottom,zhang2020global,wang2017limit} did, this paper considers the circle formation control in planar.
As shown in Fig. \ref{fig:formation_task}, a group of $N$, $N \geq 2$ fish-like robots are required to swim around a target (labeled as Robot $0$) with a radius $R_{\mathcal{C}}$ and form a preset distribution on the circle path (labeled as $\mathcal{C}$).
Considering the measurement limitations in the real world, assume that each robot can only observe the target and the two closest robots to it. Then the set of robot $i$'s two neighbors is denoted as $\mathcal{N}_i$.
For each agent $i$, $i=1,2,\ldots, N$, the distances to the target and to other robot $j$ at time $k$ are defined as $d_{i,0}(k)$ and $d_{i,j}(k)$, respectively. 
Assuming the expected distance between robot $i$ and robot $j$ is $\hat{d}_{i,j}(k)$, the goal of the circle formation control is to steer the $N$ robots to a geometry structure satisfying
\begin{eqnarray}\nonumber
	\begin{cases}
	    d_{i,0}(k) - R_\mathcal{C} = 0 \\
	    d_{i,j}(k) - \hat{d}_{i,j}(k) = 0
	\end{cases}, \; j\in \mathcal{N}_i, i = 1,2, \ldots, N
\end{eqnarray}
Note that Fig. \ref{fig:formation_task} uses an example case of $N=3$ robots just to easily represent the geometry description of the circle formation task of any number $N$.

To address this goal using DRL, a description of the circle formation task is presented from two aspects.
For one aspect, in order to make robots maintain the expected distance between each other, each robot $i$ observes the relative angle and relative distance of a nearby neighbor robot $j\in \mathcal{N}_i$.
As shown in Fig. \ref{fig:formation_task}(left), the bearing of robot $j$ relative to the orientation of robot $i$ is defined as $\phi_{i,j}(k)$, and the angle between the orientation of robot $i$ and $j$ is defined as $\alpha_{i,j}(k)$. The distance between robot $i$ and robot $j$ is $d_{i,j}(k) =||p_i(k)- p_j(k)||$,
where $p_i(k), p_j(k)$ are the position of the robot $i$ and $j$, respectively.
For another aspect, in order to make fish-like robots move on $\mathcal{C}$, each robot $i$ observes the path information relative to it including distance and angle.
As shown in Fig. \ref{fig:formation_task}(right), to obtain the signed distance $d_i(k)$ from the robot to the path, the unique projection of the robot $i$ onto the path $\mathcal{C}$ is labeled as $\textbf{P}_i$, then $d_i(k) = dist(p_i(k),\textbf{P}_i)$, where $d_i(k)$ is positive (resp. negative) when $p_i(k)$ is in (resp. outside) the circle.
Next, define the robot's current orientation $\alpha_i(k)$ relative to the path $\mathcal{C}$ with the aid of the ray $l_{\textbf{P}_i}$, where $l_{\textbf{P}_i}$ is the tangent of $\mathcal{C}$ at $\textbf{P}_i$.
Last, the motion traction point $\textbf{q}_i(k)$ is generated under the circular path perceptual region with radius $R_E$, where the traction angle $\beta_i(k)$ can be used to reflect the curvature information of the path $\mathcal{C}$.
%

% %%
% It should be mentioned that 
% \todo{we use the head orientation of each fish-like robot as the y-axis to establish an individual right-hand coordinate system for each robot to describe the circle formation task.}
% Thus, our description of the task facilitates decentralized control.

\subsection{Simulation Environment Establishment}
\begin{figure}[t]
    \centering
    \includegraphics[width=0.9\linewidth]{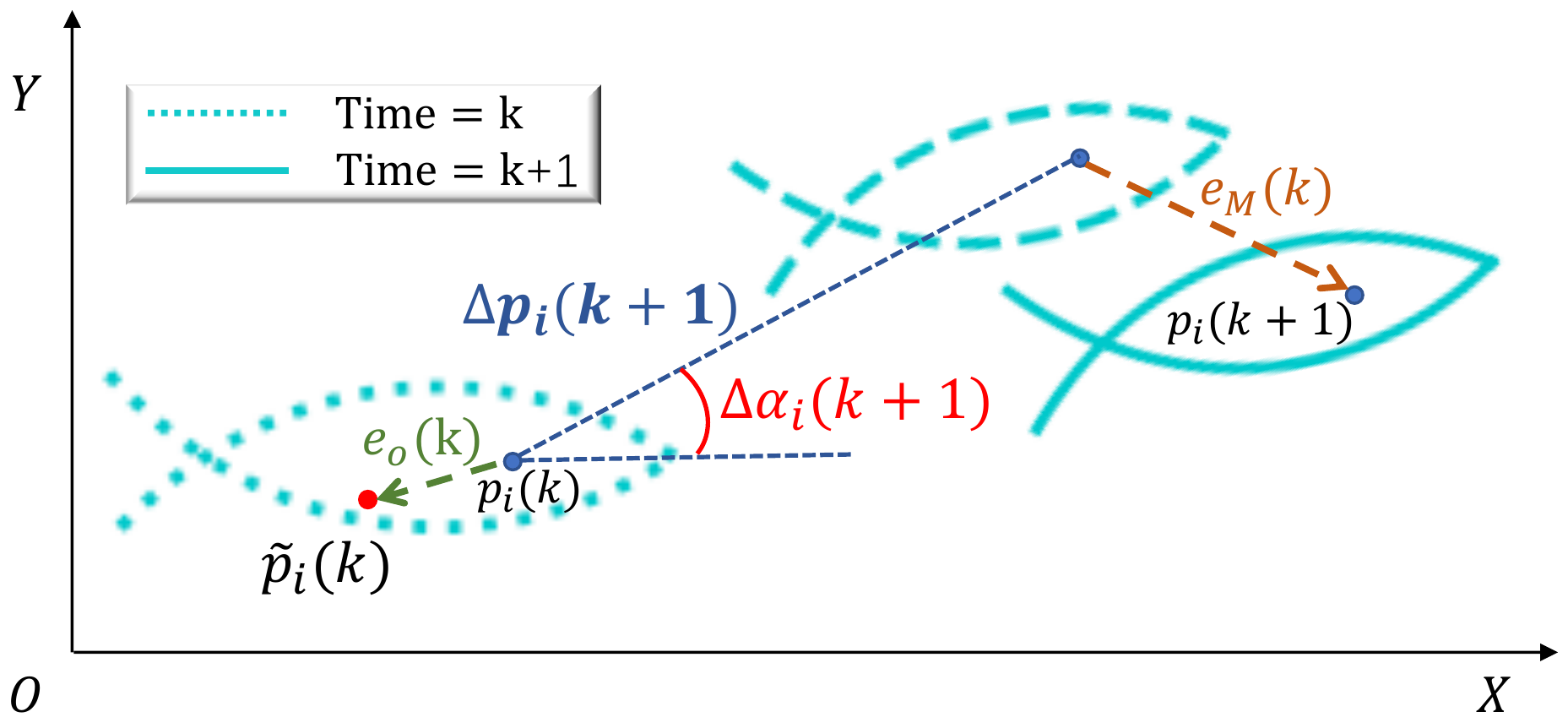}
    \caption{Simulation environment maps the CPG parameters to the kinematics of the fish-like robot with observation nosie and motion noise.} 
    \label{fig:simulation}
\end{figure}

The data-driven approach is utilized to establish the simulation environment based on the experimental motion data of the real robot to address the problem of modeling the complex dynamic model of fish-like robots, thereby reducing the gap from simulation to reality.
As shown in Fig. \ref{fig:simulation}, the simulation environment includes the data-driven based dynamic model of the fish-like robot, and the environment disturbance generator.
Specifically, some parameters of the CPG model (\ref{eq:robot_CPG}) are first fixed as $[\varphi_{12}, \varphi_{13}] = [-0.698, -2.513]rad$, $\zeta_r = 11.68/s$, $\zeta_\Phi = 5.84/s$, and changed the $R_i$, $X_i$, and $f$ to control fish-like robots to swim in the pool and collected their trajectory data.
Then, a deep neural network (DNN) is utilized with two fully-connected hidden layers of $128$ neurons to build an end-to-end mapping function $F_s$ from CPG parameters to actual motion in the real-world,
\begin{equation}
        F_s : [[R_i]_{i=1}^3(k), [X_i]_{i=1}^3(k), f] \to %[\hat{p}(k+1),\hat{\alpha}(k+1)],
        [\Delta p(k),\Delta \alpha(k)],
        \label{eq:dynamic}
    \end{equation}
where $\Delta{p}(k)\in \mathbb{R}$, and $\Delta{\alpha}(k)\in \mathbb{R}$ represent the variation of the position and that of the orientation of the fish during $t \in [k, k+1)$ (see Fig. \ref{fig:simulation}).

Next, some types of noises are added to the mapping function $F_s$ to simulate the disturbance in reality.
Considering that there are always measurement (observation) errors in reality, the observation noise $e_O$ is added in the simulation.
Assuming that the actual position of robot $i$ at time-step is $p_i(k)$, due to the observation error $e_O(k)$, the observed position becomes $\widetilde{p}_i(k)$.
The distance between agent $i$ and $j$ should be rewritten as $d_{i,j}(k) =||\widetilde{p}_i(k)- \widetilde{p}_j(k)||$.
In addition, considering that in the formation task, the water waves generated by the motion of fish-like robots will interfere with each other, the motion error $e_M(k)$ is added in the simulation, which influences the robot's position and orientation variation during $t \in [k,k+1)$.
In this way, the actual position of agent $i$ at time-step $k+1$ can be calculated by $p_i(k+1) = p_i(k)+\Delta{p}_i(k)+e_M(k)$.
In this paper, based on the experimental data, $e_O$ and $e_M$ are sampled from Gaussian distributions with the mean value of $4 cm$ and $0.5 cm$, and standard deviation of $0.5 cm$ and $0.1 cm$, respectively.
The noises are reasonable since the length of the fish-like robot is $44.3 cm$, and the maximum velocity is $50 cm/s$, where $1 s$ contains $25$ time steps.

\begin{figure}[t]
    \centering
    \includegraphics[width=0.6\linewidth]{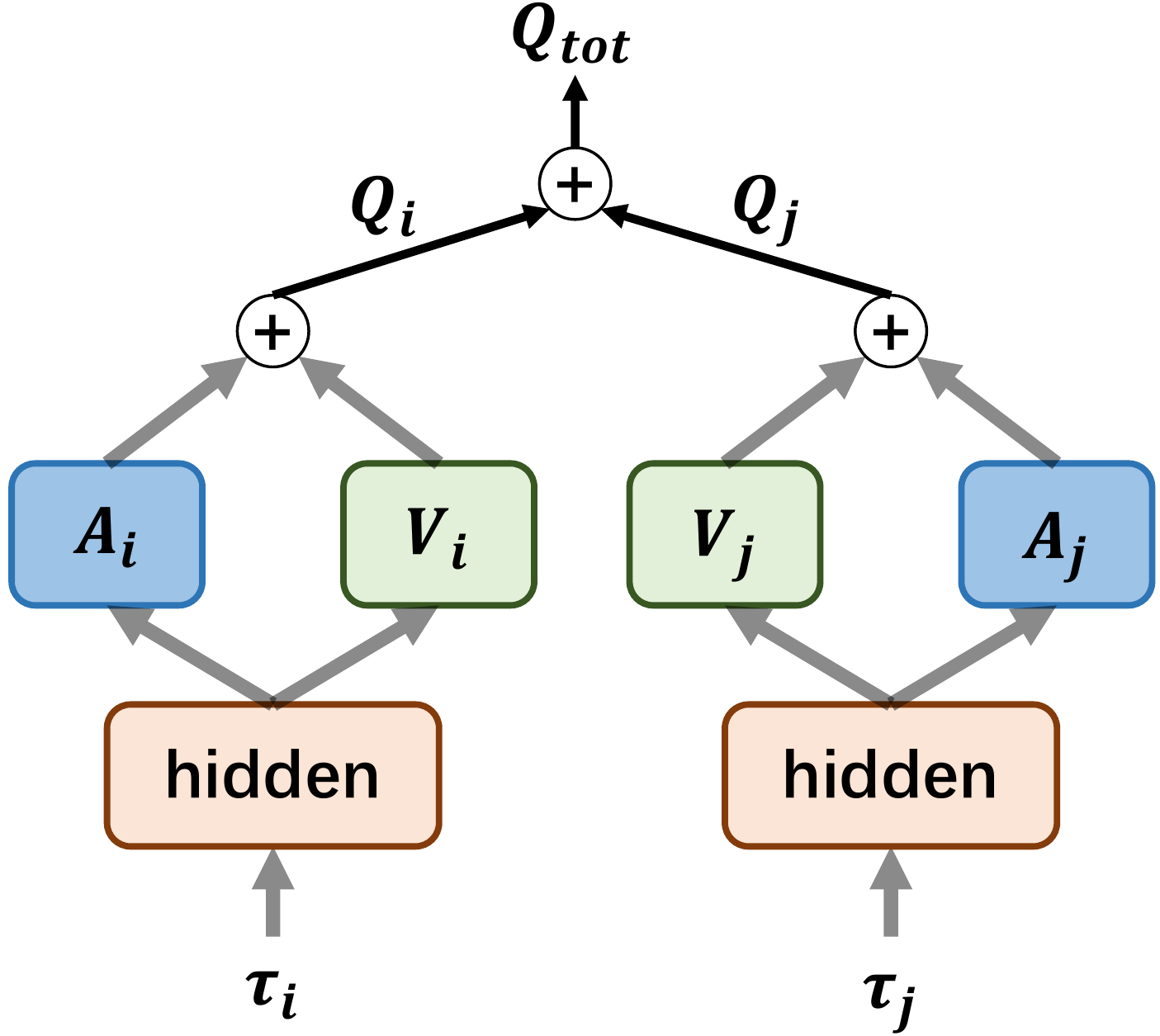}
    \caption{The framework of C2VDN, where the team Q-value $Q_\textrm{tot}$ is the sum of all individual Q-values. Each agent has a dueling network to estimate its individual Q-value, where the state value network (in green) shares parameters with others as cognitive consistency.}
    \label{fig:C2VDN}
\end{figure}

\begin{table*}[htb]
    \caption{The 15 actions of the fish-like robot, where $\Delta{p}$($cm$) and $\Delta{\alpha}$ ($rad$) are the position variation  and orientation variation of the corresponding kinematics of the robot in a time step ($1s$ contains $25$ time steps), respectively.}
    \label{tab:action}
    \centering
    \begin{tabular}{c|cc|c|cc}
        \toprule
         $\Delta{p}$~/~$\Delta{\alpha}$&\multicolumn{2}{c|}{\tabincell{c}{Turning left \\ sharply ~~~~~~~~ gradually}} & \tabincell{c}{Swimming \\straight} & \multicolumn{2}{|c}{\tabincell{c}{Turning right\\ gradually  ~~~~~~~~ sharply}}\\
        \hline
        \tabincell{c}{Low speed} &0.383/-0.0171 & 0.430/-0.0082 & 0.423/0.0 & 0.421/0.0110 & 0.382/0.0181 \\
        %\hline
        \tabincell{c}{Middle  speed}&0.748/-0.0323 & 0.851/-0.0172 & 0.856/0.0 & 0.817/0.02105 & 0.738/0.0295 \\
        %\hline
       \tabincell{c}{High speed}  &1.230/-0.0484 & 1.330/-0.0308 & 1.670/0.0 & 1.323/0.0336 & 1.194/0.0480 \\
        \bottomrule
    \end{tabular}
\end{table*}

\subsection{Cognitive Consistency-based MARL Method} 
Cooperative MARL aiming to instill in agents policies that maximize the team reward accumulated over time has achieved success in complex multi-agent tasks \cite{hernandez2019survey}.
MADDPG \cite{Ryan2017Multi} is a representative MARL method which extends DDPG \cite{LiTpHjPa2015} method to multi-agent settings by using a centralized critic for each agent.
The centralized critic network takes as input the joint actions and joint states information to approximate the action-value more properly.
After MADDPG, CTDE becomes a common paradigm for MARL (see Section \ref{sec:CTDE}).
However, the complexity the joint action-value function grows exponentially with the number of agents.

%However, training with the global information may lead to huge joint state-action space, exponentially growing with the number of agents, and thus suffer from the scalability limitation.
%
To efficiently handle this problem and achieve scalability with respect to the group size of the robots, value function factorization methods have been attached more and more attention recently.
Three representative examples of value function factorization methods include VDN \cite{2017Value}, QMIX \cite{2018QMIX}, and QTRAN \cite{2019QTRAN}.
However, the structure of VDN is too simple to have good performance, while QMIX and QTRAN could be computationally intractable in large scale number of agents task.

Therefore, we intend to design a new MARL algorithm that achieves scalability like VDN while improving its performance.
Considering that consistent cognition is crucial for good cooperation since people usually seek to have consistent cognition to the environment \cite{mcguire1966current}, we propose a new MARL method based on VDN for formation control called C2VDN which realizes the cognitive consistency of agents by parameter sharing.
Fig. \ref{fig:C2VDN} shows the overall framework of C2VDN.
Specifically, the dueling network structure is utilized to represent the action-value $Q_i$ of each agent $i$ through its state value $V_i$ and action advantage $A_i$.
Then, the state value networks of all agents which represent the cognition of the environment are sharing parameters to enable cognitive consistency.
Last, similar to VDN, the individual Q-values are summed to represent the Q-value of the team as  \vspace{-0.4cm}
\begin{equation}
    Q_\textrm{tot} = \sum_i^N Q_i = \sum_i^N (A_i + V_i).
\end{equation}

It should be mentioned that although VDN has many types of network structures, it does not take into account cognitive consistency, which is the innovation of C2VDN.
In this way, the Q-value of each agent can be updated by minimize the td-error of $Q_\textrm{tot}$,
\begin{equation}
    J_{Q_{tot}}=\mathbb{E}_{(\tau_t,u_t)\sim D}\big[Q_\textrm{tot}(\tau_t,u_t)-\hat{Q}_\textrm{tot}(\tau_t,u_t)\big]^2,
\end{equation}
where $D$ represents the replay buffer, and $\hat{Q}_\textrm{tot}(\tau_t,u_t)$ is the TD-target of the team Q-value as follows,
\begin{equation}
    \hat{Q}_\textrm{tot}(\tau_t,u_t)=r(\tau_t,u_t)+\gamma\mathbb{E}_{\tau_{t+1},u_{t+1}}[Q_\textrm{tot}(\tau_{t+1},u_{t+1})],
\end{equation}
where $\tau_{t+1}\sim D,u_{t+1}\sim \pi(u_{t+1}|\tau_{t+1})$.
$\pi(u|\tau)$ is the strategy of the team which is the combination of the strategy of each agent $\pi_i(u_i|\tau_i)$, that is, $\pi(u|\tau)=[\pi_i(u_i|\tau_i)]_{i=1}^N$.
And each agent $i$'s strategy is as follows,
\begin{equation}
    \pi_i(u_i|\tau_i) = \underset{u_i}{\arg\max}Q_i(\tau_i,u_i).
\end{equation}

\subsection{Observation, Action, and Reward}

% 邻居关系放前面了
With the basis of the description of the circle formation task (in Section \ref{sec:CFT}), especially the measurement limitations of the robots that each robot can only observe its two chosed neighbors, the observation, action, and reward are designed for MARL as follows.

\noindent\textbf{Observation:}\quad Without loss of generality, the observation of robot $1$ at time step $k$ is $\tau_1(k)=[\alpha_1(k)$, $\beta_1(k)$, $d_{1}(k)$, $d_{1,2}(k)$, $d_{1,3}(k)$, $\alpha_{1,2}(k)$, $\alpha_{1,3}(k)$, $\phi_{1,2}(k)$, $\phi_{1,3}(k)$, $\hat{d}_{1,2}(k)$, $\hat{d}_{1,3}(k)]$, where $\hat{d}_{1,2}(k)$ and $\hat{d}_{1,3}(k)$ are the expected distance to another two neighbor agents.\\
\noindent\textbf{Action:}\quad Based on the Eq. \eqref{eq:dynamic}, for simplify, we selected $15$ sets of CPG parameters as $15$ discrete actions, in which the robot has $6$ left-turn modes, $3$ straight modes, and $6$ right-turn modes. The corresponding kinematics of the fish-like robots is shown in Tab. \ref{tab:action}.

\noindent\textbf{Reward:}\quad At time step $k$, each robot $i$ will get a reward related to current formation and location to encourage better behavior. 
Taking robot $1$ as example, the object is to keep the expected distances to others while moving on the expected formation circle path. 
Therefore, the reward $r_1(k)$ is divised, consisting of two parts, $r_1(k)= r_1^d + r_1^f$. $r_1^d=-abs(d_1(k))$ represents the distance to circle path, and $r_1^f=-abs(d_{1,2}(k)-\hat{d}_{1,2}(k))-abs(d_{1,3}(k)-\hat{d}_{1,3}(k))$ represents the difference between the current formation shape and the expected formation shape.

\section{Experimental Evaluation}
The MARL strategy is trained in the simulation environment and tested both in the simulation and the real-world.
% In this section, we will describe in detail the training process, simulation experiments results , and real-world experiments results .

\begin{figure}[t]
    \centering
    \includegraphics[width=0.70\linewidth]{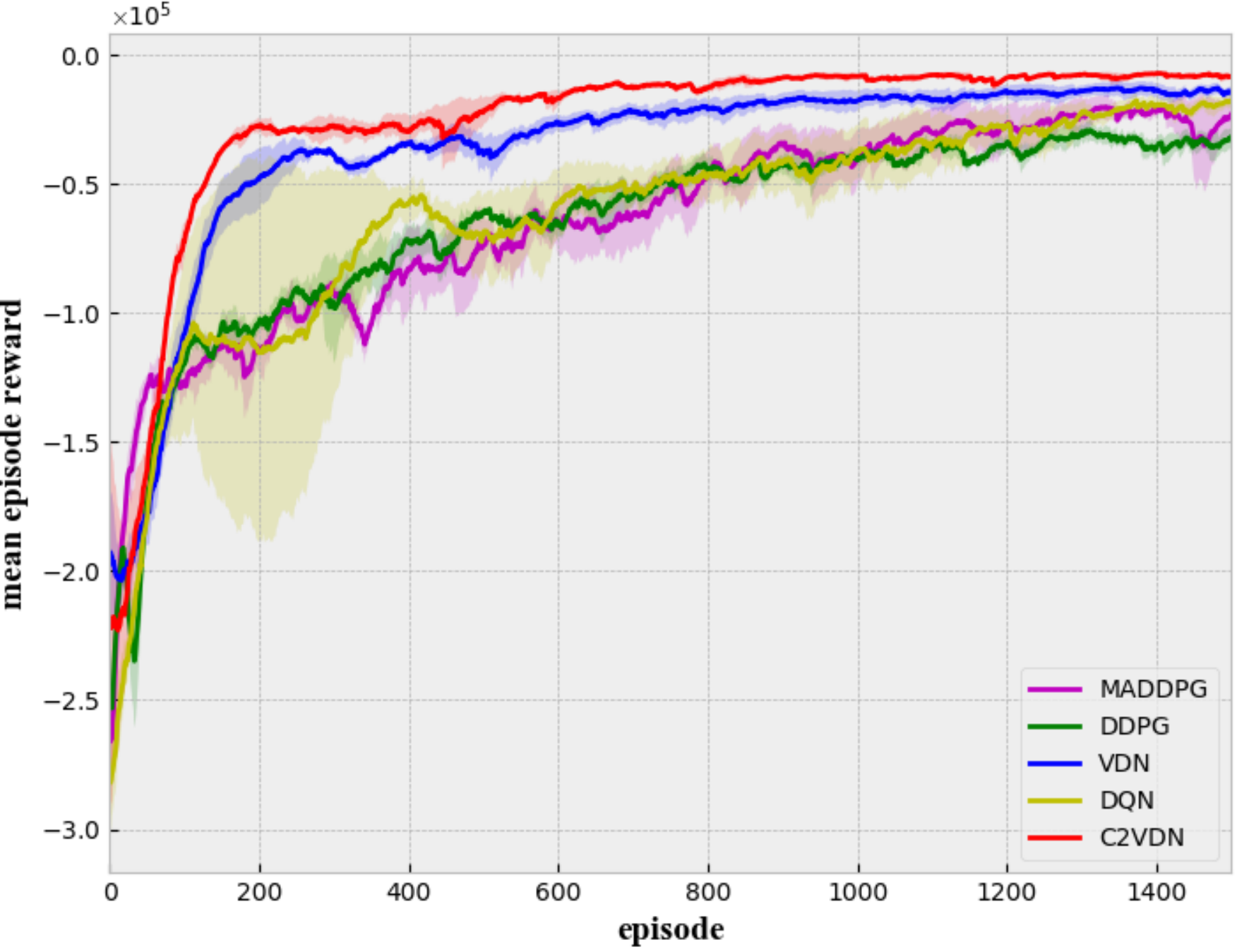}
    \caption{The mean episode reward in the training step. C2VDN outperforms DQN, DDPG, MADDPG, and VDN both in terms of convergence speed and the final performance.}
    \label{fig:reward}
\end{figure}

\subsection{Training process}\label{sec:training}
The training loop proceeds as follows.
Within the simulation environment we designed, it starts a training episode where agents are initially placed near the random circle path, that is, $R_\mathcal{C} \sim$ $Unif(60,90)$, $d_i\sim$ $Unif(-20,20)$, and $\alpha_i\sim$ $Unif(-0.2\pi,0.2\pi)$.
Then, RL strategies map current observations to robots' actions and are updated based on rewards.
Each episode runs $300$ time steps.
The RL method is trained until convergence, then save the RL model which is considered as the RL-based circle formation controller.
In order to evaluate the quality of C2VDN, C2VDN is compared with some representative RL baselines: DQN, DDPG, MADDPG, and VDN.
All RL methods are trained five times with the same setting, where the individual strategy in all methods is parameterized by a two-layer fully connected network with $64$ units per layer, and the learning rate is $0.0003$.
Fig. \ref{fig:reward} shows the performance of the five algorithms for the task during $1500$ episodes of training.

\begin{figure*}[t]
    \centering
    \includegraphics[width=0.75\linewidth]{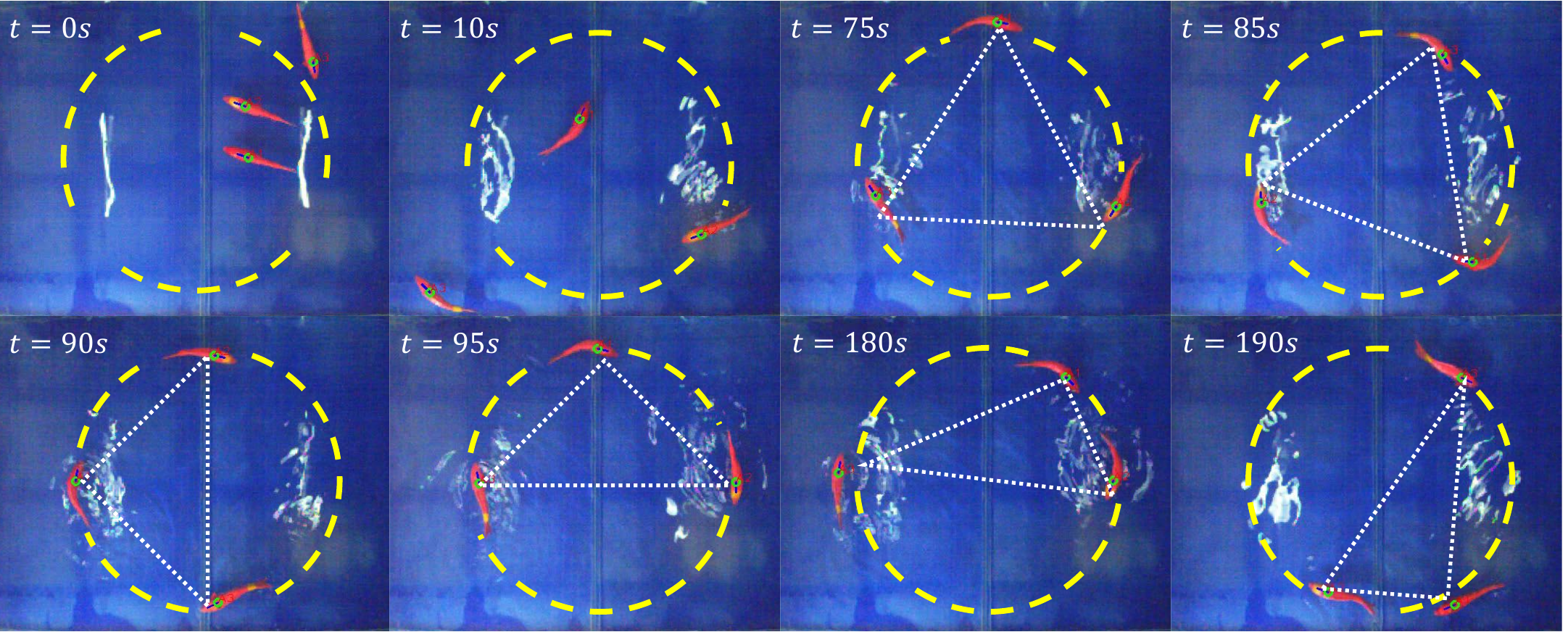}
    \caption{Snapshots of circle formation task implied by the physical platform. A group of three fish-like robots starts a circle formation task at $t=0 s$. At around $t=75 s$, they swam on a common expected circle path and presented an equilateral triangle distribution. At around $t=75 s$, the distribution was changed to an isosceles right triangle. At around $t=180 s$, the distribution was changed to a right triangle with $[1/2,1/3,1/6]\pi$ angle.}
    \label{fig:experiment_result}
\end{figure*}

\subsection{Scalability Experiments in Simulation}
In this section, the scalability experiments are conducted to evaluate whether RL-based controllers have the ability to control a large scale number of robots.
Specifically, strategies trained in three agents task are directly applied on four or ten agents to control them to form a square or regular decagonal on the circle path with $R_\mathcal{C}=80cm$.
Each agent can only observe the two agent closest to it.
To evaluate the performance, the tracking error $err_t$ and the formation error $err_f$ are defined as follows,
\begin{equation}
\begin{split}
        &err_t=\frac{1}{NT}\sum_{i=1}^N\sum_{K=1}^T(d_{i,0}(k)-R_\mathcal{C})^2,\\
        &err_f= \frac{1}{2NT}\sum_{i=1}^N\sum_{K=1}^T
    \sum_{j\in \mathcal{N}_i}(d_{i,j}(k) - \hat{d}_{i,j}(k))^2,
\end{split}
\end{equation}
where $j \in \mathcal{N}_i$ is the neighbor of agent $i$ and $|\mathcal{N}_i|=2$. 
Table \ref{tab:scalability} compares the tracking error $err_t$ and the formation error $err_f$ of the five algorithms. 
The results show that C2VDN outperforms all baselines in the ten robots task.
The results comparison between multi- and single-agent methods show that single-agent RL algorithm works little in multi-agent tasks.
The results comparison between C2VDN, VDN, and MADDPG show that C2VDN and VDN perform better than MADDPG in scalability.
The results comparison between C2VDN and VDN show that the cognitive consistency is helpful for performance improvement in the circle formation task.
The four and ten robots circle formation control experiments are intuitively shown in the supplementary video.

%
% Thus, we think C2VDN has achieved our original intention of hoping to get an RL method which is not only like VDN in scalability but also has a better performance than VDN.

% Single-agent methods (DQN, DDPG) perform poorly (especially the $err_f$) in the scalability experiments since the environment becomes more and more unstable as the number of agents increases, which cannot be considered by the single-agent method.
% %
% Thus, the excellent performance of single-RL methods during training may be caused by overfitting.
% %
% Similarly, for multi-agent methods, the difficulty of estimating the Q-values in joint action space increases as the number of agents increases.
% Thus, although the performance of MADDPG is the best in the four robots task, it is significantly bad in the ten robots task, which is due to the limitation of its algorithm mentioned in Section \ref{sec:intro}.
% %
% Both C2VDN and VDN perform well in this scalability experiment because they reduce the joint action space through value decomposition.
% %
% Same as we guessed, the C2VDN method with cognitive consistency is better than VDN both in $err_t$ or $err_f$.
% %
% Thus, we think C2VDN has achieved our original intention of hoping to get an RL method which is not only like VDN in scalability but also has a better performance than VDN.

\begin{table}[t]\hspace{-0.5cm}\vspace{-0.5cm}
    \caption{The scalability experiments results.}
    \label{tab:scalability}
    \centering
    \scriptsize
    \begin{tabular}{l|cc|cc}
    \toprule
         &  \multicolumn{2}{c|}{\tabincell{c}{ 4 robots \\ $err_t(cm)$ ~~~~~~~~ $err_f(cm)$}} & \multicolumn{2}{c}{\tabincell{c}{ 10 robots \\ $err_t(cm)$ ~~~~~~~~ $err_f(cm)$}}  \\
         \hline
    \scriptsize{C2VDN} & {$0.61\pm0.58$} & {$0.91\pm0.80$} & {$1.71\pm0.93$} & {$5.63 \pm1.00$} \\
    \scriptsize{VDN} &$0.79\pm0.25$ &$1.31\pm0.46$ & $2.57  \pm1.02$& $ 5.68 \pm 0.70 $ \\
    \scriptsize{MADDPG} &$ 0.72 \pm 0.30 $ &$0.88 \pm0.41  $ & $  4.33\pm5.59  $& $5.87 \pm 2.31 $\\
    \scriptsize{DQN} &$1.94\pm1.58$ &$2.06\pm2.34$ & $ 3.69 \pm  2.49  $& $ 7.33 \pm  3.07$\\
    \scriptsize{DDPG} &$4.47\pm 3.72$ & $5.52\pm 4.27$&$ 7.65 \pm 5.22 $ & $16.94  \pm 11.29 $\\
    \bottomrule
    \end{tabular}
\end{table}

%VDN没有一致性，所以在后面的多鱼泛化实验中效果比我们的差一些

% \begin{table}[htp]
%     \centering
%     \begin{tabular}{c|ccccc}%{p{1.3cm}|p{1.5cm}p{1.5cm}p{1.5cm}p{1.5cm}p{1.5cm}}
%          \toprule
%           & MADDPG & DDPG & VDN & DQN & OUR\\
%          \hline
%          $err_t(cm)$ & \dierhao{$ 0.72 \pm 0.30 $} & $4.47\pm 3.72$ & $0.79\pm0.25$ & $1.94\pm1.58$ &\best{$0.61\pm0.58$}\\
%          $err_f(cm)$ & \best{$0.88 \pm0.41  $} & \dierhao{$5.52\pm 4.27$} & $1.31\pm0.46$ & $2.06\pm2.34$ & {$0.91\pm0.80$}\\
%          \bottomrule
%     \end{tabular}
%     \caption{4}
%     \label{tab:simulation_results}\vspace{-0.6cm}
% \end{table}
% \begin{table}[htp]
%     \centering
%     \begin{tabular}{c|ccccc}%{p{1.5cm}|p{2cm}p{2cm}p{2cm}p{2cm}p{2cm}}
%          \toprule
%           & MADDPG & DDPG & VDN & DQN & C2VDN\\
%          \hline
%          $err_t(cm)$ & \dierhao{$  4.33\pm5.59  $} & $ 7.65 \pm 5.22 $ & $2.57  \pm1.02$ & $ 3.69 \pm  2.49  $ &\best{$1.71 \pm0.93  $}\\
%          $err_f(cm)$ & \best{$5.87 \pm 2.31 $} & \dierhao{$16.94  \pm 11.29  $} & $ 5.68 \pm 0.70 $ & $ 7.33 \pm  3.07$ & {$5.63 \pm1.00  $}\\
%          \bottomrule
%     \end{tabular}
%     \caption{10}
%     \label{tab:simulation_results}\vspace{-0.6cm}
% \end{table}

\subsection{Real World Evaluation Performance}
After the experiments in the previous subsection, C2VDN, VDN, and MADDPG are tested in the real-world.
The physical platform is shown in Fig. \ref{fig:robot_and_platform}C, which is widely used for fish-like robots research \cite{LiWaXi15,WaXiWaCa11}.
The computer processes the image flow captured by the camera and obtains relevant information on the fish-like robot, and sends the control signals to the fish-like robot through the wireless communication module.
Since the pool is only $200 cm \times 300 cm$, a circle with $70 cm$ radius was chosen as the path.
The strategy trained in simulation was directly deployed on the fish-like robots without any manual tuning.
Initially, three fish-like robots were placed randomly.
Then, they were controlled to swim on the expected common circle path and form the equilateral triangle, isosceles right triangle, and a right triangle with $[1/2,1/3,1/6]\pi$ angle in turn.

The supplementary video shows the excellent performance of our approach, and some snapshots are shown in
Fig. \ref{fig:experiment_result}.
The $err_t$ and $err_f$ of the real-world experiments are evaluated and compared with simulation results under the same settings.
Results are shown in Tab. \ref{tab: sim_to_real}.
It is obvious that C2VDN outperforms others in the real-world, which may be due to its robustness. In response to this, a more detailed study will be conducted in the future.
Considering that the length of the fish-like robot is $44.3 cm$, which is much larger than $err_t$ and $err_f$, our approach realizes the transformation from the simulation to the real world, and the strategy trained in the simulation can work excellently in the physical environment without any tuning.
Results prove the effectiveness of our RL-based approach for circle formation control.

\begin{table}[t]
    \caption{The comparison between the simulation (Sim) and physical (Real) results.}
    \label{tab: sim_to_real}
    \centering
    \begin{tabular}{c|ccc}
    \toprule
     Sim/Real & C2VDN & VDN & MADDPG \\
    \hline
         $err_t(cm)$& 0.65/2.35  & 0.64/2.62 & 0.61/2.47           \\
         $err_f(cm)$& 0.88/9.14  & 1.31/13.54 & 1.17/10.10\\
    \bottomrule
    \end{tabular}
\end{table}

\section{CONCLUSIONS}
This paper pioneered the end-to-end circle formation control of physical fish-like robots based on reinforcement learning. In the future, we will study more formation control problems in three-dimensional space and apply our approach to other robots like AUVs because our model-free approach is extremely scalable.
% 我们首创了使用强化学习方法对真实机器鱼进行端到端环形编队的方式。


%
\bibliographystyle{IEEEtran}        % Include this if you use bibtex
% \bibliography{/Users/Chen/===Working===/ref_Chen}
\bibliography{ref_Formation.bib}

\end{document}